\documentclass[a4paper]{article}

\usepackage{INTERSPEECH2019}
\usepackage{pgfplots}
\pgfplotsset{compat=1.14} 
\usepackage{graphicx, subfigure}
\usepackage{xcolor}
\definecolor{yaleblue}{rgb}{0.06, 0.3, 0.57}
\definecolor{redncs}{rgb}{0.93, 0.11, 0.14}

\title{Scaling Up Online Speech Recognition Using ConvNets}
\name{Vineel Pratap, Qiantong Xu, Jacob Kahn, Gilad Avidov, Tatiana Likhomanenko, Awni Hannun, Vitaliy Liptchinsky, Gabriel Synnaeve, Ronan Collobert}
\address{Facebook AI Research}
\email{}

\begin{document}

\maketitle
\begin{abstract}
We design an online end-to-end speech recognition system based on Time-Depth Separable (TDS) convolutions and Connectionist Temporal Classification (CTC). We improve the core TDS architecture in order to limit the future context and hence reduce latency while maintaining accuracy. The system has almost three times the throughput of a well tuned hybrid ASR baseline while also having lower latency and a better word error rate. Also important to the efficiency of the recognizer is our highly optimized beam search decoder. To show the impact of our design choices, we analyze throughput, latency, accuracy, and discuss how these metrics can be tuned based on the user requirements.

\end{abstract}
\noindent\textbf{Index Terms}: online speech recognition, low latency

\section{Introduction}
\label{sec:intro}

The process of transcribing speech in real-time from an input audio stream is known as online speech recognition. Most automatic speech recognition (ASR) research focuses on improving accuracy without the constraint of performing recognition in real time. For certain applications such as live video captioning or on-device transcription, however, low latency speech recognition is essential. In these cases, online speech recognition with a limited time delay is needed to provide a good user experience. 

Furthermore, deployment of a real-time speech recognition system at scale poses a number of challenges. First, the system must be computationally efficient to minimize the required hardware resources and to handle a large number of concurrent requests. Second, the system needs to minimize the time between a spoken word appearing in the audio and the corresponding text produced by the system. Third, the deployed recognizer should be competitive in accuracy with an offline research-grade system. We measure the three aforementioned constraints using the following metrics: (i) throughput, (ii) Real-Time Factor (RTF) and latency, and (iii) Word Error Rate (WER). 

We discuss the design of a novel ASR system that achieves 3x better throughput compared to a strong hybrid baseline. We use Time-Depth Separable convolutions~\cite{hannun2019TDS} as the building block for our acoustic model and an efficient beam-search decoder provided by the wav2letter++\footnote{https://github.com/facebookresearch/wav2letter} open-source ASR toolkit~\cite{pratap2018wav2letter}.

The rest of the paper is organized as follows. Section \ref{Sec:relatedwork} presents related work
on low latency online ASR. In Section~\ref{Sec:techdetails}, we review the design principles of our system, including acoustic models, the decoder and present novel ideas to reduce latency and greatly improve computational efficiency.In Section~\ref{Sec:setup} we discuss our experimental setup: benchmarks for measuring throughput and accuracy, as well as hardware configuration. In Section~\ref{Sec:results} we present out experimental results and ablative analysis. Finally, in Section~\ref{sec:discussion} we discuss future work and conclude.

\section{Related Work}
\label{Sec:relatedwork}

Taking an ASR system from research to a low-latency, high-throughput deployment while maintaining low WER involves non-trivial changes to the implementation. For example, many research systems use bi-directional RNNs~\cite{park2019specaug, zeyer2018improved} or Transformers~\cite{vaswani2017attention} which uses (self-)attention~\cite{chiu2018state, karita2019comparative, wang2019transformer, synnaeve2019end} which require unlimited future context and make low-latency deployment impossible. Fully convolutional acoustic models can be very efficient to train, but are often used with large future context sizes~\cite{zeghidour2018fully}. Here, we build on a large body of work in making research-grade ASR systems work in resource constrained low-latency environments.

Our architecture uses Time-Depth Separable convolution (TDS) ~\cite{hannun2019TDS} as the core building block. Bi-directional RNNs and Transformers either require considerable changes for low-latency deployment~\cite{amodei2016deep, zhang2016highway} or degrade rapidly when limiting the amount of future context~\cite{wang2019transformer}. While latency controlled bi-directional LSTMs are commonly used in online speech recognition~\cite{xue2017improving}, incorporating future context with convolutions can yield more accurate and lower latency models~\cite{peddinti2017low}. We find that the TDS convolution can maintain low WERs with a limited amount of future context.

Some recent work in low-latency speech recognition uses the RNN Transducer (RNN-T)~\cite{graves2012sequence} as the core architecture~\cite{jain2019rnn, he2019streaming}. Unlike attention models, the RNN-T decoder is causal and can be easily streamed since it does not rely on future context. Departing from prior work, we find that a CTC trained model can yield better WER than an RNN-T model while being simpler and more efficient to deploy.

Depth-wise separable convolutions~\cite{chollet2017xception} have been used in domains such as computer vision~\cite{howard2017mobilenets} and machine translation~\cite{kaiser2017depthwise} yielding dramatic reductions in model size and computational throughput while maintaining accuracy. The TDS architecture we use here also results in much lighter weight yet still low WER models. Other architectures, such as the Time-Channel Separable convolution have also shown similar gains in computational efficiency with little if any hit to accuracy~\cite{kriman2019quartznet}.

\section{Technical Details}
\label{Sec:techdetails}

In this section we describe the architectural design, algorithmic choices and various performance optimizations of our speech recognizer.


\subsection{Low-latency acoustic models}
\label{sec:lowlatency}
Our acoustic models are based on Time-Depth Separable (TDS) convolutions~\cite{hannun2019TDS}.  The reasons for choosing these models is two-fold. First, TDS blocks make use of grouped convolutions which dramatically reduce the number of parameters while still achieving low WER. This makes inference computationally efficient and keeps the model size small. Second, limiting the future context for convolution operations in TDS, which is necessary for maintaining low latency, leads to a small degradation in WER.


Figure \ref{fig:tds_block} shows the architecture of the TDS Block used in our work. The TDS block we use is modified from the original TDS implementation to make the architecture streaming friendly. The changes are described below.

\subsubsection{Asymmetrically padded convolutions}

The latency of 1-D convolutions is impacted by the number of future inputs needed to generate the current output. In order to minimize this, we use asymmetric padding for convolutions which adds more (zero) padding at the start of the input. This reduces the dependency of the current output on the future input and hence reduces the latency of the model.

For example, consider a TDS(10, 9, 80, 4) block (See Fig.~\ref{fig:tds_block} for notation) which uses symmetric padding. In order to generate the first output frame, the 1-D convolution with a filter size of 9 needs 5 input frames, since 4 frames will be padded to the start of the input. We can reduce this future context dependency to just 2 frames if we use a TDS(10, 9, 80, 1) block which pads with 8 frames at the start of the input.

\subsubsection{Removing time dependency of layernorm}

The TDS Block in~\cite{hannun2019TDS} performs layer normalization across all axes including time for a given sample. This makes the output depend on the full sample which makes streaming impractical. We resolve this by performing standard layer normalization~\cite{ba2016layer} which normalizes across the width (w) and channels (C) axes. 
For an input $x$ of shape $T \times w \ast c$, we compute the layernorm output $\hat{x}$ as
\begin{equation}
 \hat{x}[i,j] = g * \frac{x[i,j] - \mu_i}{ \sqrt{\sigma_i ^ 2 + \epsilon}} + b  \quad i \in [0, T),  j \in [0, w * c)
\end{equation}
where $\mu_i = \frac{1}{w * c} \sum_{j} x[i,j] $, $\sigma^2_i =  \frac{1}{w * c} \sum_{j} (x[i,j] - \mu_i)^2 $, $g$ and $b$ are scalar affine transform parameters and $\epsilon$ is a small constant added for numerical stability.



\begin{figure}[!ht]
\centering     
\subfigure[]{\label{fig:tds_block}\includegraphics[width=25mm]{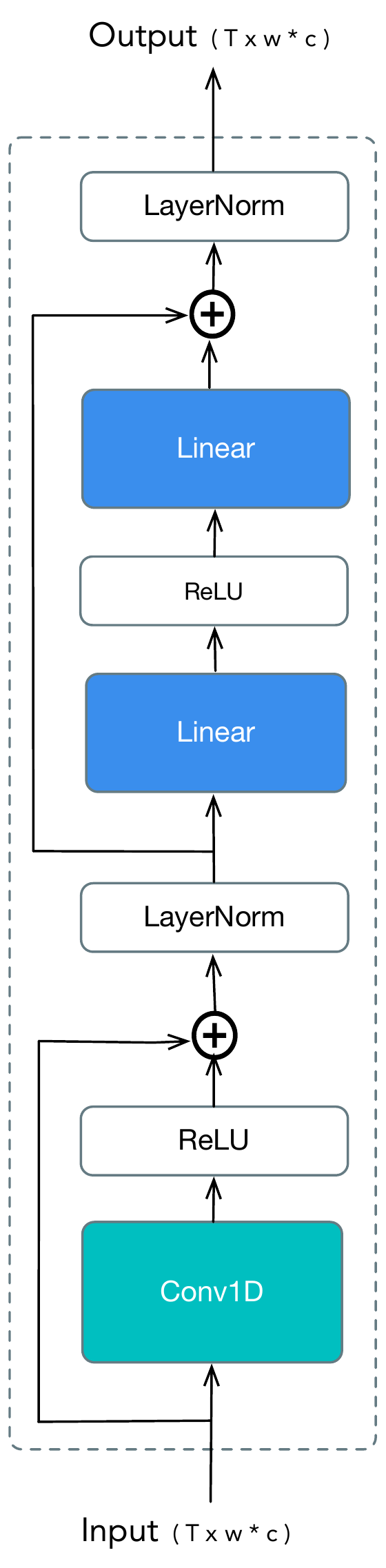}}
\hspace{6mm}
\subfigure[]{\label{fig:encoder}\includegraphics[width=24mm]{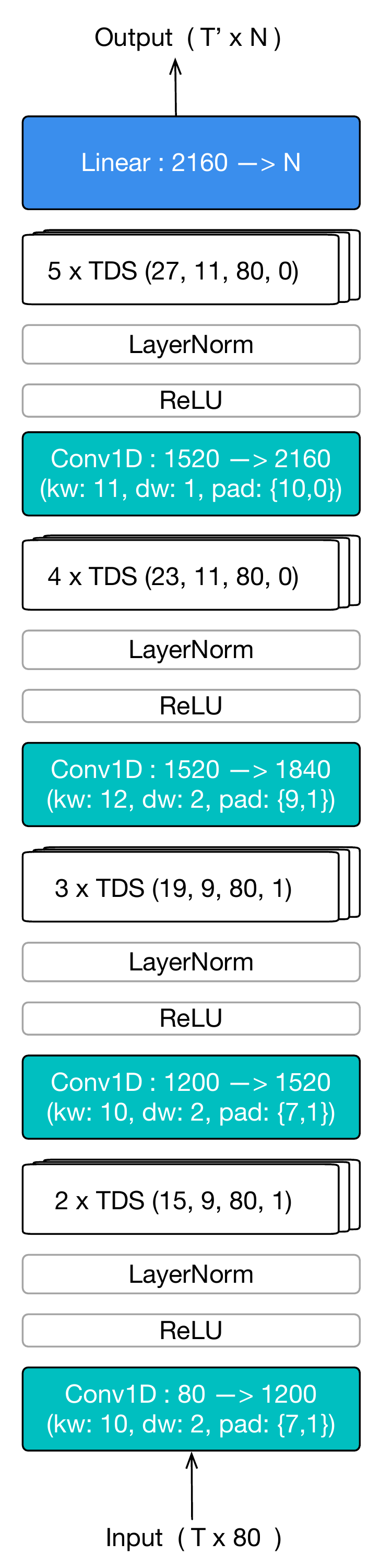}}
\caption{(a) Time-Depth Separable Convolutional Block, TDS($c$, $kw$, $w$, $rPad$). Parameters of Conv1D: in,out channels=$w*c$, stride=$1$, filter\_size=$kw$, num\_groups=$w$, padding=$\{kw - 1 - rPad, rPad\}$. Parameters of Linear: in,out channels=$w*c$ (b) Acoustic Model Architecture. Notation: $dw$ = stride, $kw$ = filter size}
\end{figure}

Figure \ref{fig:encoder} shows the architecture of our acoustic model. The model consists of two 15-channel, three 19-channel, four 23-channel and five 27-channel TDS blocks. They are separated by 1-D convolution layers which increase the number of output channels and optionally perform subsampling. A final linear layer produces an N-dimensional output (N is the size of the token set) which is later passed to a log-softmax layer prior to computing the CTC loss. The model has a total of 104 million parameters and has a total subsampling factor of 8. It takes 80-dimensional log mel-scale filter bank features as input. The features are extracted with a stride of 10ms so the model has a receptive field of $\sim$10 seconds per frame and a future context of 250ms.

\subsection{Online Beam Search Decoding} 

To integrate a language model, we develop an online version of the beam search decoder based on wav2letter++~\cite{pratap2018wav2letter}. The online decoder consumes the encoded frames for the current input audio chunk and extends the beam search graph. We output the most likely sequence of words based on this extended beam search graph for every chunk. Since the best path can change as we extend the beam search graph, we allow for correction of partial transcriptions generated earlier. Also, in order not to have the history in memory grow infinitely with incoming audio chunks, pruning is applied to the history buffer after consuming a certain number of chunks.  

Apart from the existing performance optimizations included in wav2letter++ decoder \cite{zeghidour2018fully}, we introduce two further pruning techniques. First, we consider only top-K (e.g. $K = 50$) tokens according to the acoustic model score, when expanding each hypothesis in the beam. This is also commonly known as acoustic pruning. Second, we propose only the blank symbol if its posterior probability is larger than 0.95 \cite{park2018fully}. With these optimizations as well as the 8x reduction in the number frames from subsampling in the acoustic model, we find that decoding time accounts for about 5\% of the overall inference procedure.

\subsection{Inference Implementation} 
\label{inference_impl}
We wrote a standalone, modular platform to perform the full online inference procedure. The pipeline takes audio chunks as input and processes them in an online manner. To perform the matrix multiplications and 1-D group convolutions required by the acoustic model, we use the 16-bit floating point FBGEMM\footnote{https://github.com/pytorch/FBGEMM} implementation. We have carefully optimized memory use by relying on efficient I/O buffers and extensively profiled to remove any excess computational overhead.

\section{Experimental Setup}
\label{Sec:setup}
\subsection{Data}
The training set used for our experiments consists of around 1 million utterances ($\sim$13.7K hours) of in-house English videos publicly shared by users.  The data is completely anonymized and doesn't contain any personally-identifiable information (PII). We report results on two test sets - vid-clean and vid-noisy consisting of around 1.4K utterances each ($\sim$20 hours). More information about the data set can be found in \cite{le2019senones}.

\subsection{Training}
 All our experiments are run using wav2letter++ framework \cite{pratap2018wav2letter} with 64 GPUs for each experiment. We use 80-dimensional log mel-scale filter banks as input features, with STFTs computed on 25ms Hamming windows strided by 10ms. For the output token set, we use a vocabulary of 5000 sub-word tokens generated from SentencePiece toolkit \cite{kudo2018subword}. We use local mean and variance normalization instead of global normalization on the input features prior to the acoustic model so that the system can run in an online manner. Local normalization computes the summary statistics over the prior $n$ frames and uses them to normalizing the input features at the current frame.  We use $n=300$ for all of our experiments which corresponds to a window size of approximately $3$ seconds. We also use SpecAugment \cite{park2019specaug} for data augmentation in all of the experiments.

\subsection{Baseline Systems}
\label{Subsec:baselines}
We compare our work with two strong baselines. All the systems (including ours) use the same training, validation and test sets.

\noindent\textbf{Baseline 1} \quad
The first baseline~\cite{le2019senones} is a hybrid system based on context-dependent graphemes (chenones) and uses multi-layer Latency Controlled Bidirectional Long Short-Term Memory layers (LC-BLSTM)\cite{Zhang_2016} in the acoustic model. The system is initially bootstrapped with cross entropy (CE) training and then fine-tuned with the lattice-free maximum mutual information (LF-MMI) \cite{povey2016purely} loss function. 

\noindent\textbf{Baseline 2} \quad
The second system~\cite{jain2019rnn} is based on the RNN-T \cite{graves2012sequence} architecture and uses Latency Controlled Bidirectional Long Short-Term Memory layers (LC-BLSTM)~\cite{Zhang_2016} in the acoustic model. The token set consists of 200 sentence pieces, constructed with the sentence piece library~\cite{kudo2018subword}. Unlike Baseline 1, the second baseline is trained in an end-to-end manner. 

\subsection{Evaluation Benchmarks}
For consistency in results, all the benchmarks, including baselines, are run on Intel Skylake CPUs with 18 physical cores and 64GB of RAM. We describe the performance metrics evaluated in our experiments below 

\subsubsection{Real-Time Factor}
Real-Time Factor (RTF) is the ratio between the time taken to process the input and the input
duration. For a system to be considered real-time, RTF should be $\le$ 1. RTF is dependent on the number of concurrent streams being run by the system. Further, we define ``RTF@40" as the RTF using 40 concurrent streams.

\subsubsection{Throughput}
Throughput is defined as the rate at which audio is consumed. In other words, it is the  number of audio seconds processed per wall clock second.

\subsubsection{User-perceived latency}
\label{sec:e2e_latency}
As the name suggests, user-perceived latency is the latency metric most relevant to the end user. It is affected by chunk size, RTF, acoustic model context, and decoder output delay. Quantifying latency as a function of these parameters is difficult. Instead we compute this latency in an empirical and end-to-end manner by measuring the timestamp of when a transcribed word is available to the user and compare this with the same word's timestamp in the original audio.

More formally, we define user-perceived latency as the average delay between the end timestamp of the words in the reference transcript and time when it is shown to the user by the system. As an example, consider a 1 sec audio file with transcript ``how are you" and the corresponding start and end word timestamps for the words as (100ms-200ms), (300ms-400ms), (500ms-600ms) respectively. Lets consider an ASR system with RTF 0.2 which processes the audio file and outputs the words ``how", ``are" after consuming 500ms of audio and the word ``you" after looking at 1000ms of the audio. 

For this system, word ``the” will be produced at 600ms (500ms for audio chunk to be available and 100ms (500ms * RTF) for processing it while the true timestamp of end of the word is 200ms which gives us a latency of 400ms. Computing this for every word and taking the average, we can see that this system has average user-perceived latency of (400ms + 200ms + 500ms) / 3 = 366.67 ms.

We have measured user-perceived latency on a set of 1000 samples from the TIMIT~\cite{timit} corpus, which comes with word alignments in advance. We note that all the ASR systems that we consider produce the correct transcript on these samples (WER = 0), which is necessary for this latency analysis.

\section{Results and Analysis}
\label{Sec:results}

Table \ref{tab:results} shows the performance comparison of our system with the two baselines mentioned in \ref{Subsec:baselines}. We use 750ms chunk size for all the experiments with our system. User-perceived latency is measured at 40 concurrent streams. We can see that our system is able to achieve better WER at a much higher throughput even when using 16-bit floating-point precision as compared to the 8-bit fixed point precision used by the baselines.

\begin{table}[htb]
\setlength{\tabcolsep}{4pt} 
\begin{footnotesize}
  \centering
  \begin{tabular}{lccc}
  \toprule
     & \textbf{Baseline 1} & \textbf{Baseline 2} & \textbf{Our System} \\
     & \tiny{LC-BLSTM + LF-MMI} & \tiny{LC-BLSTM + RNN-T} & \tiny{TDS + CTC} \\
  \midrule
  \midrule
  vid-clean WER & 14.1  & 13.93  & 13.19 \\
  vid-noisy WER & 22.15 & 22.58 & 21.16 \\
  \midrule
  Total Parameters & 80 mil & 60 mil & 104 mil \\
  Inference Precision & INT8 & INT8 & FP16 \\
  Throughput & 55 & 64 & 147 \\
  RTF@40  & 0.70 & 0.60 & 0.26 \\
  \begin{tabular}{@{}l@{}}User-perceived \\ latency\end{tabular} & 1.18 sec & - & 1.09 sec \\
  \bottomrule
  \end{tabular}
  \caption{A comparison of our system with the baseline systems.}
  \label{tab:results}
\end{footnotesize}
\end{table}


\subsection{Effect of low-latency acoustic models}
We performed an ablation study to observe the change in WER from the original TDS model~\cite{hannun2019TDS} with global input normalization to the low latency version we propose here. Table \ref{tab:stream_tbl} shows that there is only a $\sim$4\% relative increase in WER because of the changes we make to decrease the latency of the model. We observe that limiting the future context of the acoustic model contributes the most to the degradation in WER.

\begin{table}
\begin{small}
\setlength{\tabcolsep}{4pt} 
 \begin{tabular}{lcc}
  \toprule
     & \textbf{vid-clean}  & \textbf{vid-noisy}  \\
       & \textbf{WER} & \textbf{WER} \\
  \midrule
  Original TDS with input globalnorm & 12.64 & 20.45 \\
  $\quad$ + globalnorm $\rightarrow$ localnorm & 12.72 & 20.46 \\
  $\quad$ + remove time-axis for layernorm & 12.71 & 20.44 \\
  $\quad$ + 250ms future context limit & 13.19 & 21.16 \\
  \bottomrule
  \end{tabular}
  \caption{The effect on WER when decreasing the latency of the TDS model}
  \label{tab:stream_tbl}
 \end{small}
\end{table}

\begin{table}
\centering
\begin{small}
  \centering
  \begin{tabular}{lcc}
  \toprule
    \textbf{Future context} & \textbf{vid-clean WER}  & \textbf{vid-noisy WER} \\
  \midrule
  250 ms & 13.19 & 21.16 \\
  500 ms & 13.07 & 20.81 \\
  1000 ms & 12.92 & 20.46 \\
  2500 ms & 12.80 & 20.73 \\
  5000 ms & 12.65 & 20.44 \\
  \bottomrule
  \end{tabular}
  \caption{Effect of future context on WER performance}
  \label{latency}
\end{small}
\end{table}

\begin{figure}
 \centering
 \includegraphics[width=\linewidth]{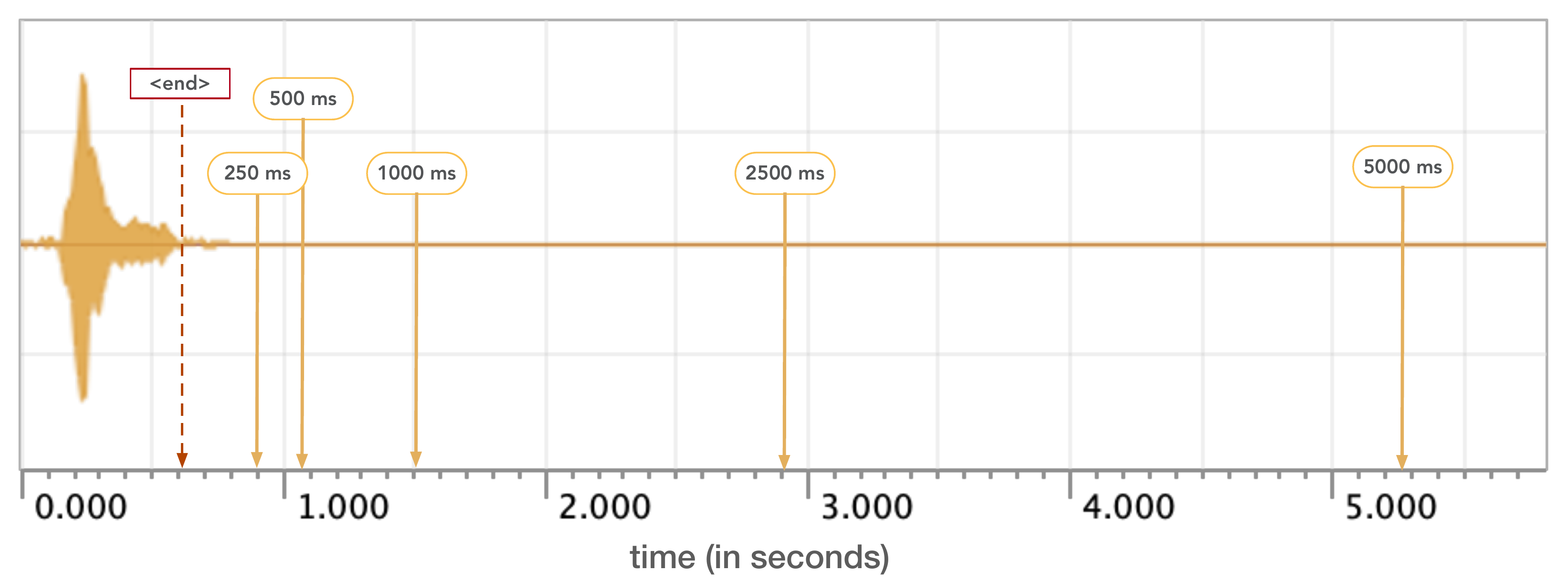}
 \caption{The acoustic model latency when varying the future context size. The transcript of input audio is ``yes" and end of this word is marked in the audio with \texttt{<end>}. Word timestamps generated by greedy decoding for different future context sizes are marked at various locations in the input audio.}
 \label{fig:timestamp}
\end{figure}

Table \ref{latency} shows the effect of WER with varying future context sizes. We have kept the receptive field of our convolutional acoustic model fixed at 10 seconds, use local normalization and remove time from layer normalization axes for all the experiments. We see that the WER almost always improves as the future context size is increases. Figure \ref{fig:timestamp} shows the word timestamps generated from greedy decoding on an input audio file with only one spoken word. We see that the transcripts can be produced faster by using smaller future context sizes.

These results demonstrate that asymmetric padding is important for low latency convolutional acoustic models. We see a significant reduction in latency with just $\sim$ 4\% relative drop in WER going from a model with symmetric convolutions (5 sec future context) to a model with asymmetric convolutions (250ms future context).

\subsection{Effect of concurrent streams}

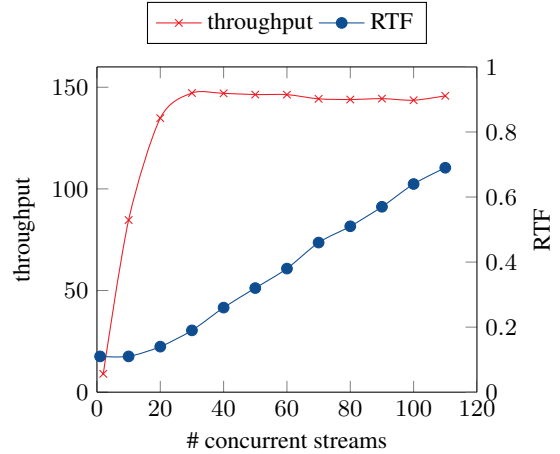
\begin{figure}
\centering
\begin{small}
\begin{tikzpicture}
\pgfplotsset{
    width=5cm,compat=1.3,
    scale only axis,
    xmin=0, xmax=120
}

\begin{axis}[
  axis y line*=left,
  ymin=0, ymax=160,
  xlabel=\# concurrent streams,
  ylabel=throughput,
  legend style={at={(0.95,0.25)},anchor=north east},
  legend columns=-1
]
\addplot[smooth,mark=x,redncs]
  coordinates{
    (2,9.02)
    (10,84.68)
    (20,134.80)
    (30,147.16)
    (40,146.99)
    (50,146.40)
    (60,146.36)
    (70,144.30)
    (80,143.96)
    (90,144.40)
    (100,143.61)
    (110,145.78)
}; \label{plot_one}
\end{axis}
\begin{axis}[
  axis y line*=right,
  axis x line=none,
  ymin=0, ymax=1,
  ylabel=RTF,
  legend style={at={(0.5,1.2)},anchor=north},
  legend columns=-1
]
\addlegendimage{/pgfplots/refstyle=plot_one}\addlegendentry{throughput}
\addplot[smooth,mark=*,yaleblue]
  coordinates{
    (1,0.11)
    (10,0.11)
    (20,0.14)
    (30,0.19)
    (40,0.26)
    (50,0.32)
    (60,0.38)
    (70,0.46)
    (80,0.51)
    (90,0.57)
    (100,0.64)
    (110,0.69)
}; \addlegendentry{RTF}
\end{axis}
\end{tikzpicture}
\end{small}
\caption{RTF, throughput trade-off vs concurrent streams}
\label{fig:rtfplof}
\end{figure}

As discussed in Section \ref{sec:intro}, processing multiple audio streams in parallel is necessary to achieve higher throughput. Figure \ref{fig:rtfplof} shows the effect of increasing the number of concurrent stream on throughput and RTF for a fixed chunk size of 750ms. We see that throughput increases dramatically up to about 30 concurrent streams beyond which it no longer improves. On the other hand, RTF consistently increases as we increase the number of concurrent streams. While any setting with RTF $<$ 1 is considered real-time, as discussed in Section \ref{sec:e2e_latency}, lower RTF improves the user-perceived latency.  

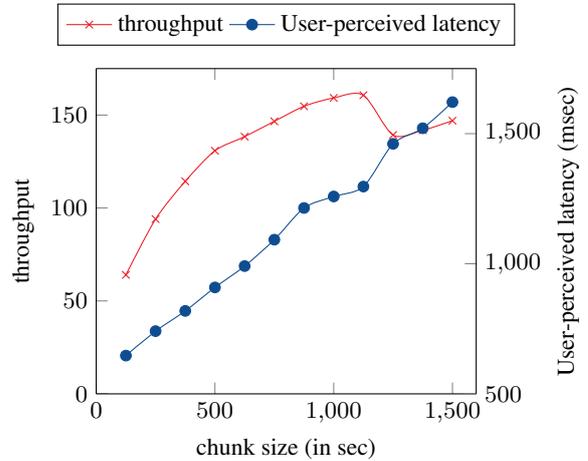
\begin{figure}
\begin{small}
\begin{tikzpicture}
\pgfplotsset{
    width=5cm,compat=1.3,
    scale only axis,
    xmin=0, xmax=1600
}

\begin{axis}[
  axis y line*=left,
  ymin=0, ymax=175,
  xlabel=chunk size (in sec),
  ylabel=throughput,
  legend style={at={(0.5,1.2)},anchor=north},
  legend columns=-1
]
\addplot[smooth,mark=x,redncs]
  coordinates{
    (125, 64.02454376)
    (250, 94.02082825)
    (375, 114.3546295)
    (500, 130.8069763)
    (625, 138.3946838)
    (750, 146.6342621)
    (875, 154.7318115)
    (1000, 159.224884)
    (1125, 160.6692657)
    (1250, 139.274704)
    (1375, 141.8369141)
    (1500, 146.9654694)
}; \label{plot_two}
\end{axis}
\begin{axis}[
  axis y line*=right,
  axis x line=none,
  ymin=500, ymax=1750,
  ylabel=User-perceived latency (msec),
  legend style={at={(0.5,1.2)},anchor=north},
  legend columns=-1
]
\addlegendimage{/pgfplots/refstyle=plot_two}\addlegendentry{throughput}
\addplot[smooth,mark=*,yaleblue]
  coordinates{
    (125, 647)
    (250, 741)
    (375, 818.739)
    (500, 908.96)
    (625, 991.09)
    (750, 1092.538)
    (875, 1214.329)
    (1000, 1258.558)
    (1125, 1296.57)
    (1250, 1460.86)
    (1375, 1520.78)
    (1500, 1621.21)
}; \addlegendentry{User-perceived latency}
\end{axis}
\end{tikzpicture}
\end{small}
\caption{Latency, throughput trade-off vs chunk size}
\label{fig:chunkplot}
\end{figure}

\subsection{Effect of chunk size}
For streaming speech recognition, we feed the acoustic model with audio in chunks every $T$ milliseconds (where $T$ is the chunk size). The transcript is then generated in an online fashion. Figure~\ref{fig:chunkplot} shows the effect of increasing the audio chunk size on throughput and latency. We see that even though throughput can be increased by increasing chunk size, the user-perceived latency also increases.

\section{Conclusion and Future Work}
\label{sec:discussion}
We present a convolutional online speech recognition system which outperforms two strong baselines in throughput, WER and latency. We also demonstrate the trade-offs in improving one metric over another and how the metrics can be fine-tuned according to the application requirements. In future work we plan to further improve throughput by using 8-bit fixed precision quantization and techniques including reducing the model depth on demand~\cite{fan2019reducing} and weight sparsity~\cite{kalchbrenner2018efficient}. The above techniques may also help reduce the model size while speeding up inference time. We have also deployed our models on-device, but we leave a full analysis with competitive benchmarks to future work.

\section{Acknowledgements}
\label{sec:ack}
We would like to thank Mahaveer Jain and Duc Le for help in evaluating the baseline models.

\bibliographystyle{IEEEtran}

\bibliography{mybib}


\end{document}